%% file: root.tex
\documentclass[letterpaper, 10 pt, conference]{ieeeconf}  

\IEEEoverridecommandlockouts                              

\usepackage{epsfig} 
\usepackage{amsmath} 
\usepackage{amssymb}  
\usepackage{xspace}
\usepackage{xcolor}
\usepackage{color}
\usepackage{cite}
\usepackage{hyperref}
\usepackage{booktabs, multirow}
\usepackage{colortbl}
\usepackage{bm}
\usepackage{tikz}
\usepackage{setspace}
\PassOptionsToPackage{table}{xcolor}
\usepackage[font=footnotesize,labelfont=bf]{caption}
\newcounter{RNum}
\renewcommand{\theRNum}{\arabic{RNum}}
\newcommand{\Remark}{\noindent\textit{\textbf{Remark}~\refstepcounter{RNum}\textbf{\theRNum}: }}

\newcommand{\methodname}{SMART\xspace}

\title{\LARGE \bf
\textit{\methodname}: 
Advancing Scalable Map Priors for \\ Driving Topology Reasoning
}

\author{Junjie Ye$^{1*}$, David Paz$^2$, Hengyuan Zhang$^{3*}$, Yuliang Guo$^2$, Xinyu Huang$^2$, \\Henrik I. Christensen$^3$, Yue Wang$^1$, and Liu Ren$^2$
\thanks{$^{*}$Work done while interned at Bosch Research North America.}
\thanks{$^{1}$Thomas Lord Department of Computer Science, University of Southern California 
        {\tt\small \{yejunjie, yue.w\}@usc.edu}}%
\thanks{$^{2}$Bosch North America and Bosch Center for AI (BCAI)
        {\tt\small \{david.pazruiz, yuliang.guo2, xinyu.huang, liu.ren\}@us.bosch.com}}%
\thanks{$^{3}$Contextual Robotics Institute, UC San Diego
        {\tt\small \{hyzhang, hichristensen\}@ucsd.edu}}%
}

\begin{document}

\maketitle
\thispagestyle{empty}
\pagestyle{empty}

\begin{abstract}
Topology reasoning is crucial for autonomous driving as it enables comprehensive understanding of connectivity and relationships between lanes and traffic elements. 
While recent approaches have shown success in perceiving driving topology using vehicle-mounted sensors, their scalability is hindered by the reliance on training data captured by consistent sensor configurations.
We identify that the key factor in scalable lane perception and topology reasoning is the elimination of this sensor-dependent feature.
To address this, we propose \methodname, a scalable solution that leverages easily available standard-definition (SD) and satellite maps to learn a map prior model, supervised by large-scale geo-referenced high-definition (HD) maps independent of sensor settings. Attributed to scaled training, \methodname alone achieves superior offline lane topology understanding using only SD and satellite inputs. Extensive experiments further demonstrate that \methodname can be seamlessly integrated into any online topology reasoning methods, yielding significant improvements of up to 28\% on the OpenLane-V2 benchmark. Project page: \href{https://jay-ye.github.io/smart}{https://jay-ye.github.io/smart}.


\end{abstract}

\section{Introduction}
In recent years, lane perception and topology reasoning of driving scenes has received considerable
attention, providing essential information about the structure and connectivity of road elements for autonomous driving and driver assistance
systems. 
While previous map perception methods~\cite{li2021hdmapnet, liu2023vectormapnet, MapTR, maptrv2} primarily focus on identifying road markers, \textit{e.g.}, lane dividers, road boundaries, and pedestrian crossings, driving topology reasoning encompasses the broader understanding of not only the lane geometry but also their connectivity and relationships with traffic elements, such as traffic lights and road signs. Such in-depth topological reasoning is essential for downstream tasks such as trajectory prediction, path planning, and motion control~\cite{sadat2020eccv, Nayakanti2023icra, mao2024agentdriver}.

\begin{figure}[!t]
\centering
  \includegraphics[width=0.9\linewidth]{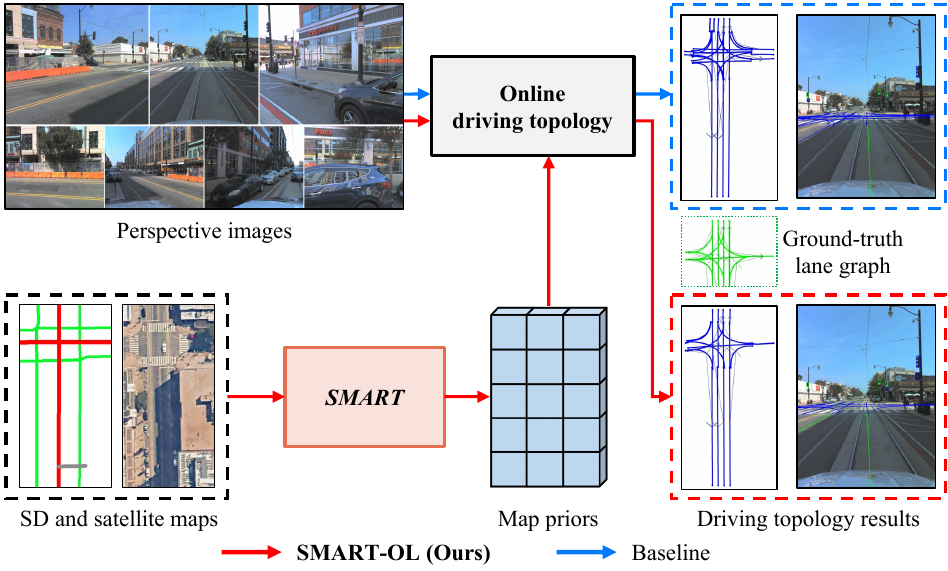}
\caption{\textbf{Comparison between baseline and \methodname-OL.}
Existing topology reasoning methods suffer from limited sensor data. \methodname augments online topology reasoning with robust map priors learned from scalable SD and satellite maps, substantially improving lane perception and topology reasoning.
}
\label{fig:teaser}
\vspace{-13pt}
\end{figure}

In the literature, driving topology is formulated as a graph that captures the location of lane centerlines and traffic elements, along with the connectivity between centerlines and their relationships to each traffic element~\cite{wang2023openlanev2}. Recent advancements~\cite{li2023toponet, wu2024topomlp, ma2024roadpainter} have shown promising progress in this field. 
Nevertheless, existing approaches face significant limitations. They are trained on data captured with consistent sensor configurations.
Scaling these models typically requires large datasets collected from vehicles with uniform sensor setups, which is both costly and time-consuming. Moreover, the low perspective of ground-based vehicles leads to occlusions from other vehicles, buildings, and objects, posing substantial challenges. Some recent works~\cite{wang2023openlanev2, luo2024smerf, zhang2024enhancingonlineroadnetwork} aim to mitigate these occlusions by integrating standard definition (SD) maps with sensor inputs to provide geometric and topological priors. Yet, despite the widespread availability of SD maps, these approaches remain constrained by the limited availability of sensor data for training.

Training models for driving topology reasoning also requires access to high-definition (HD) maps associated with sensor data. Although these models are sensor-dependent, the supervision signals--HD maps--are calibrated to the real world and are geo-referenced. Setting aside the dependency on sensor data, there exists large-scale HD maps independently of sensor sets. For example, as shown in Table~\ref{tab:statistics}, the motion forecasting set of Argoverse 2~\cite{Argoverse2} possesses HD maps from 285 times more scenes than those typically used for online mapping, not to mention HD maps available from many other sources~\cite{nuplan, Sun2020waymo}. The extensive scale of HD maps offers a huge potential for learning generalizable map priors.

On the other hand, SD maps and satellite images are universally accessible from crowd sources~\cite{OpenStreetMap, mapbox_raster_tiles_api} with geo-locations. These \textit{geospatial} maps offer powerful priors for understanding lane structures and are updated at a reasonable frequency. This leads us to an important question: \textit{Can we advance scalable online topology reasoning with easily accessible geospatial maps and large-scale HD maps?}

We address this challenge by proposing a simple yet effective two-stage driving topology reasoning pipeline.
As illustrated in Fig.~\ref{fig:teaser}, in the first stage, we introduce \textit{\methodname}, a map prior model targeted at reason lane topology using SD and satellite maps, supervised by large-scale HD maps. In the second stage, \methodname is integrated with \textit{online} topology reasoning models, enhancing them with powerful map priors learned from the first stage. 
This approach shifts the dependency on massive high-quality sensor data with consistent configurations to highly accessible and scalable geospatial maps, enabling scaled learning of adaptable map representations. 
The well-trained \methodname, in turn, provides robust priors that significantly enhance the generalizability of online topology reasoning. 
Remarkably, our map prior model alone achieves state-of-the-art \textit{offline} lane topology\footnote{We refer to generating lane graphs from \textit{offline} SD and satellite maps as \textit{offline} lane topology.} given only SD and satellite inputs. Extensive experiments further demonstrate that integrating \methodname into existing methods boosts performance by a large margin.


\begin{table}[!t]\centering
    \begin{spacing}{0.8}
    \caption{\textbf{Statistics of available HD maps in different datasets.} 
    The motion forecasting set from Argoverse (AV) 2~\cite{Argoverse2} has over 280 times more scenes compared to the sensor datasets.}\label{tab:statistics}
    \end{spacing}
    \footnotesize
    \resizebox{\linewidth}{!}{%
        \begin{tabular}{cccc}\toprule
        Datasets &Num. of scenes &Num. of HD maps \\\cmidrule{1-3}
        AV 2 sensor - $train$ &700 &22,477 \\
        nuScenes - $train$ &700 &27,968 \\\cmidrule{1-3}
        AV 2 motion forecasting - $val$ &24,988 & 549,736 \\
        AV 2 motion forecasting - $train$ &199,908 &4,397,976\\
        \bottomrule
        \end{tabular}
    }
    \vspace{-10pt}
\end{table}

        
To summarize, our contributions are three-fold:
\begin{itemize}
    \item A simple yet effective architecture for map prior learning at scale, achieving impressive lane topology reasoning with SD and satellite inputs.
    \item A map prior model that can be seamlessly integrated into any topology reasoning framework, enhancing robustness and generalizability.
    \item Evaluations on the widely-used benchmark~\cite{wang2023openlanev2} underscore the effectiveness of \methodname in driving topology reasoning, achieving state-of-the-art performance. 
\end{itemize}

\begin{figure*}[!t]
\centering
  \includegraphics[width=0.95\linewidth]{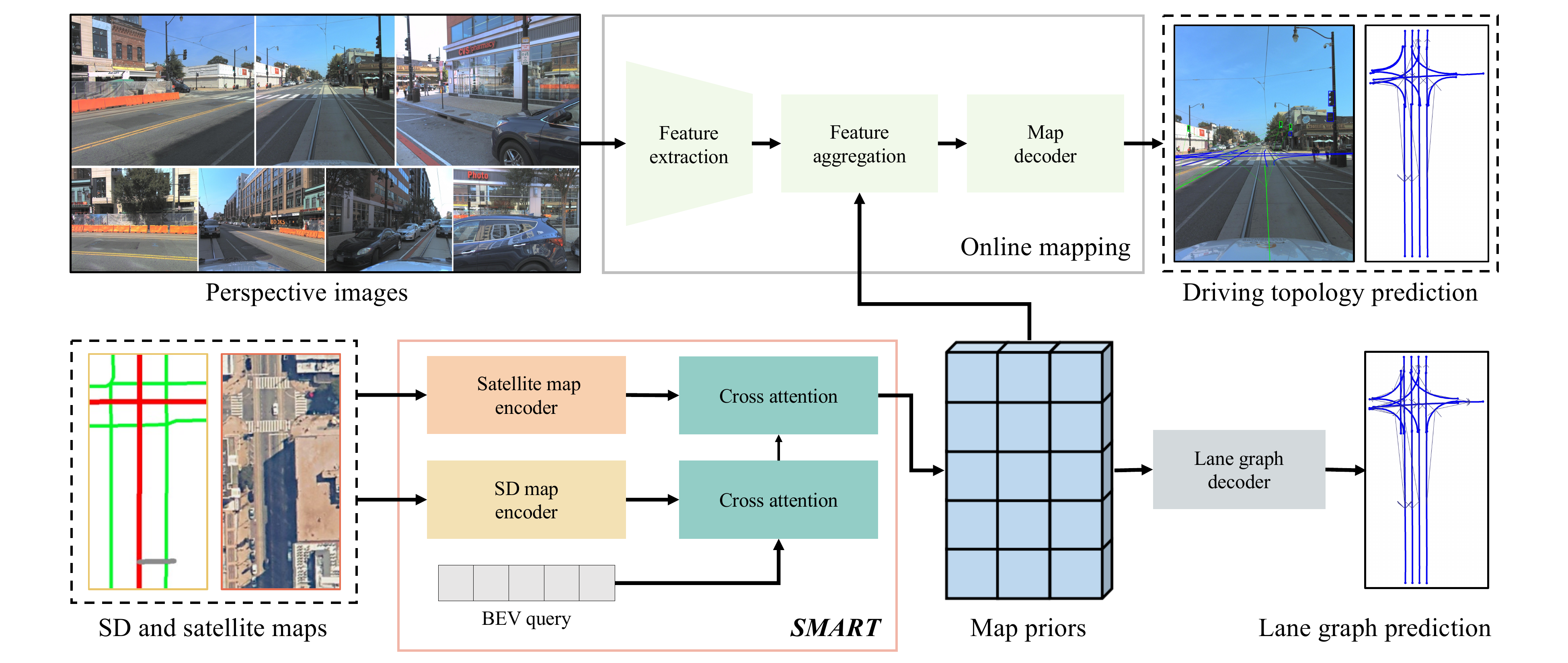}
  \vspace{-3pt}
\caption{\textbf{Outline of the proposed approach.} 
In the first stage (bottom row), \methodname is trained at scale using SD and satellite maps for lane graph prediction, supervised by large-scale geo-referenced HD maps. In the second stage (top row), the robust map priors learned by \methodname are seamlessly integrated into any online driving topology reasoning models, significantly enhancing lane perception and topology reasoning. 
}
\label{fig:outline}
\vspace{-13pt}
\end{figure*}

\section{Related Works}
\subsection{Online Mapping}

Online mapping seeks to perceive lane structures on the fly, offering instant scene information for downstream tasks, in contrast to offline methods that rely on traffic observations~\cite{zurn2024autograph} or aerial images~\cite{Buchner2023UrbanLaneGraph, Blayney_2024_CVPR_bezier}. Early works primarily focus on estimating the geometry of road elements~\cite{li2021hdmapnet, liu2023vectormapnet, MapTR, maptrv2}. As one of the pioneering works, HDMapNet~\cite{li2021hdmapnet} generates an online semantic map, followed by post-processing to obtain vectorized road elements. Liu \textit{et al.}~\cite{liu2023vectormapnet} model vectorized HD map learning in an end-to-end fashion. Liao \textit{et al.}~\cite{MapTR, maptrv2} propose a permutation invariant loss that boosts map geometry learning. Recent efforts have extended online mapping to include topology reasoning, which involves understanding the connectivity and relationships between lanes and traffic elements~\cite{wang2023openlanev2, li2023toponet, wu2024topomlp, fu2024topologic}. Graph-based approaches like TopoNet~\cite{li2023toponet} use scene graph neural networks to model these relationships, while simpler models like TopoMLP~\cite{wu2024topomlp} still achieve promising improvements in topology reasoning with multilayer perceptrons (MLPs). Other methods~\cite{li2024topo2d, fu2024topologic} explore novel ways to improve the initialization of 3D queries and interpret geometric distances for better topology reasoning. However, a common bottleneck for these methods is the reliance on large amounts of sensor data collected under consistent configurations.

\subsection{Map Priors}
Incorporating prior knowledge has been shown to advance online mapping performance~\cite{xiong2023neuralmapprior, luo2024smerf, Gao2024ICRA, zhang2024enhancingonlineroadnetwork, Jiang2024pmapnet, Yuan2024presight}. Xiong \textit{et al.}~\cite{xiong2023neuralmapprior} assume a multi-traversal setting to enhance map perception with features from previous traversals. More recently, some works~\cite{luo2024smerf, zhang2024enhancingonlineroadnetwork} integrate SD maps with surroundings view images for joint training, demonstrating improved topology modeling performance. Similarly, Gao~\textit{et al.}~\cite{Gao2024ICRA} aid road element detection with satellite imagery. In spite of promising improvements achieved, priors introduced in these works remain consistently coupled to limited sensor data, leading to unsatisfying scalability.
By contrast, we aim to learn a unified map prior representation from massive geospatial maps, featuring generalizability to novel locations and compatibility with any online topology reasoning models.
Additionally, we identify the benefits of combining the road-level topology priors from SD maps and comprehensive bird's eye view (BEV) textures from satellite maps.


\section{Methodology}
\subsection{Problem Definition}
Driving topology reasoning aims to perceive the geometric layout of lane centerlines $\mathbf{V}_{\rm l}$ and traffic elements $\mathbf{V}_{\rm t}$, and reason lane-to-lane connectivity $\mathbf{E}_{\rm ll}$ and lane to traffic element relationship $\mathbf{E}_{\rm lt}$. They formulate two graphs, \textit{i.e.}, $(\mathbf{V}{\rm _l}, \mathbf{E}_{\rm ll})$ and $(\mathbf{V}_{\rm t} \cup \mathbf{V}_{\rm l}, \mathbf{E}_{\rm lt})$. Specifically, $\mathbf{V}_{\rm l}$ consists of a set of directed lane instances with each denoted as $\mathbf{v}_{\rm l}=[p_0, ..., p_{n-1}]$, where $p=(x,y,z)\in \mathbb{R}^3$. 

Existing works estimate lane topology directly from perspective images captured by $C$ synchronized surround-view cameras mounted to the ego-vehicle, some with SD maps as additional input. This line of work lacks scalability and generalizability due to its dependence on sensor data. 

The pipeline of the proposed approach is shown in Fig.~\ref{fig:outline}. During inference, we first fetch SD and satellite maps corresponding to the desired GPS location and adopt a well-trained map prior model to extract prior features. These features are then integrated into online topology reasoning models for enhanced centerline detection and relation modeling. To tackle the limited availability of sensory data and make full use of easily accessible SD and satellite maps, we decouple prior and sensor inputs with two-stage training.

\subsection{Offline Map Prior Learning}
In the first stage, we train a map prior model, referred to as \methodname, with the goal of inferring lane graphs $(\mathbf{V}_{\rm l}, \mathbf{E}_{\rm ll})$ from offline SD and satellite maps.

\Remark We focus on lane graph modeling only in the first stage due to the invisibility of traffic elements like traffic lights and signs from the aerial perspective.
\subsubsection{SD map fetching}\label{sec:sd_fetching} We fetch SD maps from OpenStreetMap (OSM)~\cite{OpenStreetMap} as in previous work~\cite{luo2024smerf, zhang2024enhancingonlineroadnetwork}. SD maps supply comprehensive geographic information for locations worldwide. Given a specific GPS location and orientation, we extract a local SD map that contains $m$ road polylines $\mathbf{R}=[\mathbf{r}_1, ..., \mathbf{r}_{m}]$. Each polyline is represented as an ordered set of 2D points and includes $k$ associated attributes $\bm{\alpha} \in \mathbb{R}^{m \times k}$ such as road types and lane counts. The number of polylines and their individual lengths may vary. The extracted map is then rotated and cropped to generate an ego-centric view covering the desired spatial range.

\subsubsection{Satellite map fetching}\label{sec:sate_fetching} We obtain satellite maps from the Mapbox Raster Tiles API~\cite{mapbox_raster_tiles_api}, which segments the Earth's surface into a grid of tiles at various zoom levels. By projecting a given GPS coordinate onto the 2D map tile plane, we identify the appropriate tile indices and query the API to retrieve the relevant tiles. The zoom level is set to 20, corresponding to $\sim$0.11 meters per pixel. Through a process of map tile stitching, rotation, and cropping, we generate the satellite map $\mathbf{S}\in \mathbb{R}^{H\times W}$ that aligns with a specific location, orientation, and spatial range. 

\subsubsection{SD and satellite map encoding}
For SD maps, we begin by evenly sampling $N$ points along each polyline, then pad the number of polylines to $m'$ to maintain a consistent count across maps within each training batch.
This reformulates each SD map as $\mathbf{R'}=[\mathbf{r'}_1, ..., \mathbf{r'}_{m}, \mathbf{0},...]_{m'}$ and $\bm{\alpha'} \in \mathbb{R}^{m' \times k}$, where $\mathbf{r'} \in \mathbb{R}^{N \times 2}$. Following~\cite{luo2024smerf}, we transform $\mathbf{R'}$ from 2D coodinates to corresponding sinusoidal embeddings $\mathbf{R''} \in \mathbb{R}^{m' \times Nd}$ with $d$ dimensions using sinusoidal encodings $\rm E_{sin}$:
\begin{equation}
    \mathbf{R''} = {\rm E_{sin}} (\mathbf{R'}) \quad.
\end{equation}

With the refined road polylines $\mathbf{R''}$ and associated attributes $\bm{\alpha'}$, we generate SD map features $\mathbf{F}_{\rm SD} \in \mathbb{R}^{m' \times C}$ with a linear layer and a Transformer~\cite{vaswani2017attention} encoder:
\begin{equation}
    \mathbf{F}_{\rm SD} = {\rm Enc} ( {\rm Linear} ( {\rm Concat} (\mathbf{R''}, \bm{\alpha'}) ), \mathbf{M} ) \quad ,
\end{equation}
where $\mathbf{M}$ is a binary mask indicating valid polylines.

For satellite maps, we adopt an image backbone $\mathcal{F}$ to extract features as $\mathbf{F}_{\rm Sate}=\mathcal{F}(\mathbf{S})$ and flatten them to the dimension of $\mathbf{F}_{\rm Sate} \in \mathbb{R}^{H_{\rm f}W_{\rm f} \times C}$. 

To encode SD and satellite maps as a unified prior feature, we sequentially cross-attend features extracted from SD and satellite maps to a BEV feature map encoded with position embeddings, denoted as $\mathbf{B} \in \mathbb{R}^{H_{\rm B} W_{\rm B} \times C}$. Therefore, the fused prior features $\mathbf{B}_{\rm prior}$ can be obtained as: 
\begin{equation}
    \begin{split}
    \hat{\mathbf{B}} =& {\rm Softmax}(\mathbf{B} \mathbf{F}_{\rm SD}^{\top})\mathbf{F}_{\rm SD} \\
    \mathbf{B}_{\rm prior} =& {\rm Softmax}(\hat{\mathbf{B}} \mathbf{F}_{\rm Sate}^{\top})\mathbf{F}_{\rm Sate}
    \end{split} \quad .
\end{equation}

\subsubsection{Offline lane graph decoding}

To decode the lane graph from the prior features $\mathbf{B}_{\rm prior}$, we begin by attending them to learnable centerline instance queries $\mathbf{Q}\in \mathbb{R}^{N_{\rm L} \times C}$ with decoder layers in deformable DETR~\cite{zhu2021deformable} that combine self-attention, deformable attention, and feed-forward network. Sequentially, we adopt the simplified GCN in~\cite{li2023toponet} to modulate $\mathbf{Q}$ for enhanced relational modeling. Hence, the enhanced instance queries $\hat{\mathbf{Q}}$ is formulated as:
\begin{equation}
    \hat{\mathbf{Q}} = {\rm GCN} ( {\rm Dec} (\mathbf{Q}, \mathbf{B}_{\rm prior}) ) \quad .
\end{equation}

Next, two sets of three-layer MLPs are adopted to classify and regress $N_{\rm L}$ lanes, each with $N_{\rm P}$ points as follows:
\begin{equation}
    \mathbf{c} ={\rm MLP_{cls}}(\hat{\mathbf{Q}}) \quad\quad \mathbf{P} ={\rm MLP_{reg}}(\hat{\mathbf{Q}}) \quad ,
\end{equation}
where $\mathbf{c}\in \mathbb{R}^{N_{\rm L}}$ are the confidence scores and $\mathbf{P}\in \mathbb{R}^{N_{\rm L} \times (N_{\rm P}\times 3)}$ are regressed lane sets.

On the other hand, to achieve topological reasoning, the instance query $\hat{\mathbf{Q}}$ is fed into 2 MLPs separately, resulting in $\hat{\mathbf{Q}}'_1$ and $\hat{\mathbf{Q}}'_2$, both with a shape of $N_{\rm L} \times \frac{C}{2}$. These two features are then repeated along a new axis and concatenate together as $\Theta_{\rm ll} \in \mathbb{R}^{N_{\rm L} \times N_{\rm L} \times C}$. Subsequently, a binary classifier is operated on $\Theta_{\rm ll}$ to obtain the final topological matrix $\bm \epsilon$.

Consequently, the lane graph $(\mathbf{V}_{\rm l}, \mathbf{E}_{\rm ll})$ is predicted by further filtering out low-confidence centerlines in $\mathbf{P}$ and $\bm \epsilon$.

\subsubsection{Learning objective} Similar to existing works~\cite{li2023toponet, wu2024topomlp, ma2024roadpainter}, the overall loss $\mathcal{L}$ consisting of classification loss $\mathcal{L}_{\rm cls}$, regression loss $\mathcal{L}_{\rm reg}$, and topological loss $\mathcal{L}_{\rm top}$ is defined as:
\begin{equation}
    \mathcal{L} = \mathcal{L}_{\rm cls} + \mathcal{L}_{\rm reg} + \mathcal{L}_{\rm top} \quad ,
\end{equation}
where $\mathcal{L}_{\rm cls}$ and  $\mathcal{L}_{\rm top}$ employ focal loss, $\mathcal{L}_{\rm reg}$ adopts L1 loss.



\subsection{Online Topology Reasoning with \methodname}\label{sec:online_integration}

The well-trained \methodname can be seamlessly integrated into any online topology reasoning models, augmenting them with robust prior features derived from SD and satellite maps. Generally, there are two mainstream pipeline used in existing online approaches: BEV-based and perspective-based methods. The former~\cite{li2023toponet, fu2024topologic, ma2024roadpainter} explicitly projects features extracted from surrounding views into the BEV perspective using BEVFormer~\cite{li2022bevformer}, followed by perception and reasoning heads for topology prediction. In this pipeline, we directly substitute the learnable BEV queries with prior features extracted by \methodname. Conversely, in perspective-based methods~\cite{wu2024topomlp, li2024topo2d}, the lane decoder interacts directly with multi-view visual features for perception and topology reasoning. In this paradigm, we employ a cross-attention layer to align prior features with perspective features. 

To mitigate the risk of overfitting due to the limited sensor data, the weights of \methodname are kept fixed during the training of online topology models, which are trained with their original settings. We term online topology reasoning with \methodname integrated as \textit{\methodname-OL}.

\Remark In practical applications, aside from fetching and inferring prior features online, these features can be precomputed using future geo-locations of the ego-vehicle derived from the navigation route or mission plans.



\section{Experiments}


In this section, we aim to answer the following questions: (1)~How far can we get with \methodname alone? (2)~How much can \methodname boost online topology reasoning?
(3)~How well can \methodname generalize to unseen areas? (4)~Can \methodname benefit from scaled training data? (5)~Can SD and satellite map fusion boost performance? (6)~Can \methodname reduce the reliance on sensor data?

\subsection{Dataset and Metrics}

\subsubsection{Dataset} Without loss of generality, we utilize the Argoverse 2 motion forecasting dataset~\cite{Argoverse2} for training \methodname, which provides geo-referenced HD maps despite lacking associated sensor data. Ground-truth lane graphs are derived from these HD maps by regressing centerlines and connecting lanes based on topology~\cite{wang2023openlanev2}. 
Table~\ref{tab:statistics} summarizes the number of scenes in different datasets, each comprising consecutive driving frames at 2 Hz, with each frame associated with an HD map segment. While traditional online mapping benchmarks like the Argoverse 2 sensor dataset~\cite{Argoverse2} and nuScenes~\cite{caesar_nuscenes_2020} provide only 700 driving scenes with fewer than 30k HD maps, the validation set of the motion forecasting dataset includes $\sim$25k scenes with over 549k HD maps. The training set contains $\sim$\textbf{200k} scenes and over \textbf{4.3 million} HD maps--\textbf{285} times more scenes and over \textbf{160} times more HD data compared to sensor sets. This substantial increase in data amount and diversity supports the learning of significantly more generalizable and scalable map priors.

\begin{table}[!b]\centering
\vspace{-10pt}
    \begin{spacing}{0.8}
    \caption{\textbf{Comparison of \methodname and online mapping methods on lane graph generation.} Per-frame latency is measured by running individual methods on an Nvidia GeForce RTX 3090 GPU. Benefiting from scaled training, \methodname outperforms \textit{online} mapping methods on both lane perception and topology reasoning with \textit{offline} geospatial maps.
    }\label{tab:direct_hd_mapping}
    \end{spacing}
    \footnotesize
    \resizebox{\linewidth}{!}{%
    \begin{tabular}{cccccc}\toprule
        Input type &Methods &${\rm DET_l}$ &${\rm TOP_{ll}}$ &Latency (ms) \\\cmidrule{1-5}
        \multirow{2}{*}{Perspective images} 
        &TopoNet &28.6 &10.9 & 172.6 \\ 
        &TopoMLP &28.5 &21.7 & 328.2 \\ 
        \cmidrule{1-5}
        SD and satellite maps &\methodname (Ours) &\textbf{37.9} &\textbf{31.9} &\textbf{44.3} \\
        \bottomrule
    \end{tabular}
    }
\end{table}

\input{tab/online_comparison}

The evaluation is conducted on the primary set of the OpenLane-V2 dataset~\cite{wang2023openlanev2}, which extends Argoverse 2 sensor set~\cite{Argoverse2} with ground truth for traffic element detection and topology relationship association. The associated 7-view images are only available during training of online mapping.

We extract the corresponding ego-centric SD maps from OSM~\cite{OpenStreetMap} and satellite images from Mapbox Raster Tiles~\cite{mapbox_raster_tiles_api} using the methods described in Sections~\ref{sec:sd_fetching} and~\ref{sec:sate_fetching}. All maps are centered on the ego-vehicle's position and oriented in the direction of travel, covering an area of $100{\rm m} \times 50{\rm m}$ along the longitudinal and lateral directions, respectively. 



\subsubsection{Metrics} 
We report the OpenLane-V2 Score (OLS) defined in~\cite{wang2023openlanev2}, which is computed as:
\begin{equation}
    \text{OLS} = \frac{1}{4} \left [ \text{DET}_{\rm l} + \text{DET}_{\rm t} + \sqrt{\text{TOP}_{\rm ll}} + \sqrt{\text{TOP}_{\rm lt}} \right ] \quad ,
\end{equation}
where DET$_{\rm l}$ is the discrete Fréchet distance~\cite{eiter1994frechetdistance} mean average precision (mAP) for lane geometry, DET$_{\rm t}$ is the mAP for traffic elements. TOP$_{\rm ll}$ and TOP$_{\rm lt}$ are the topology scores for lane-to-lane connectivity and lane-to-traffic-element relationship, respectively.

\subsection{Implementation details}
We implement \methodname with PyTorch. The number of sampling points $N$ in SD maps is set to 11. We adopt a 6-layer Transformer encoder as the SD map encoder. For satellite map encoding, we first resize satellite images to $500\times 250$ and then utilize ResNet-50~\cite{He2016resnet} pretrained on ImageNet~\cite{Krizhevsky2012imagenet} as the backbone for features extraction. Multi-scale features from the last three stages of ResNet-50 are employed and cross-attented with a BEV feature map $\mathbf{B}$, which has a size of $H_{\rm B} = 200$, $W_{\rm B} = 100$, and $C=256$. The number of centerline queries $N_{\rm L}$ is set to 200. To keep the training time manageable while maintaining the diversity of HD maps, we sample frames in each scene with a frame rate of 0.5 and filter out still frames. This results in 200k scenes with 810k HD maps, which are used to train \methodname in the first stage, unless otherwise specified. \methodname is trained for 8 epochs with AdamW~\cite{adamw} optimizer on 8 NVIDIA A100 GPUs, which takes $\sim$2 days to complete training. The batch size for each GPU is 12. 

We integrate well-trained \methodname into two state-of-the-art open-sourced baselines for second-stage training, namely TopoNet~\cite{li2023toponet} and TopoMLP~\cite{wu2024topomlp}, which represent the BEV-based and perspective-based models, respectively. We maintain their default setting of using ResNet-50 as the image backbone for multi-view images. All other settings remain unchanged, except for the integration of map prior features from \methodname, as described in Sec.~\ref{sec:online_integration}.

\subsection{How far can we get with \methodname alone?}\label{sec:direct}

\methodname is trained to generate lane graphs solely from geospatial maps in the first stage. Naturally, this raises the question of how accurate the lane graphs predicted from SD and satellite maps are. In Table~\ref{tab:direct_hd_mapping}, we compare \methodname with two state-of-the-art \textit{online} driving topology reasoning methods~\cite{li2023toponet, wu2024topomlp} on the validation set of~\cite{wang2023openlanev2}. Metrics related to traffic elements are excluded due to the impracticality of detecting them from an aerial view. The results show that \methodname achieves superior performance, with ${\rm DET_l}$ of 37.9 compared to 28.6 and 28.5 for TopoNet and TopoMLP. In terms of ${\rm TOP_{ll}}$, \methodname scores 31.9, significantly higher than TopoNet's 10.9 and TopoMLP's 21.7. Additionally, \methodname shows significantly lower per-frame latency of 44.3 ms, which is $3.9\times$ and $7.4\times$ faster than online mapping models.
Despite differences in input modality and the amount of training data, the remarkable performance of \methodname underscores the abundant prior information contained in geospatial maps and the vast potential for scaling driving topology reasoning from a geospatial map perspective.



\subsection{How much can \methodname boost online topology reasoning?}

\begin{figure}[!t]
\centering
  \includegraphics[width=0.99\linewidth]{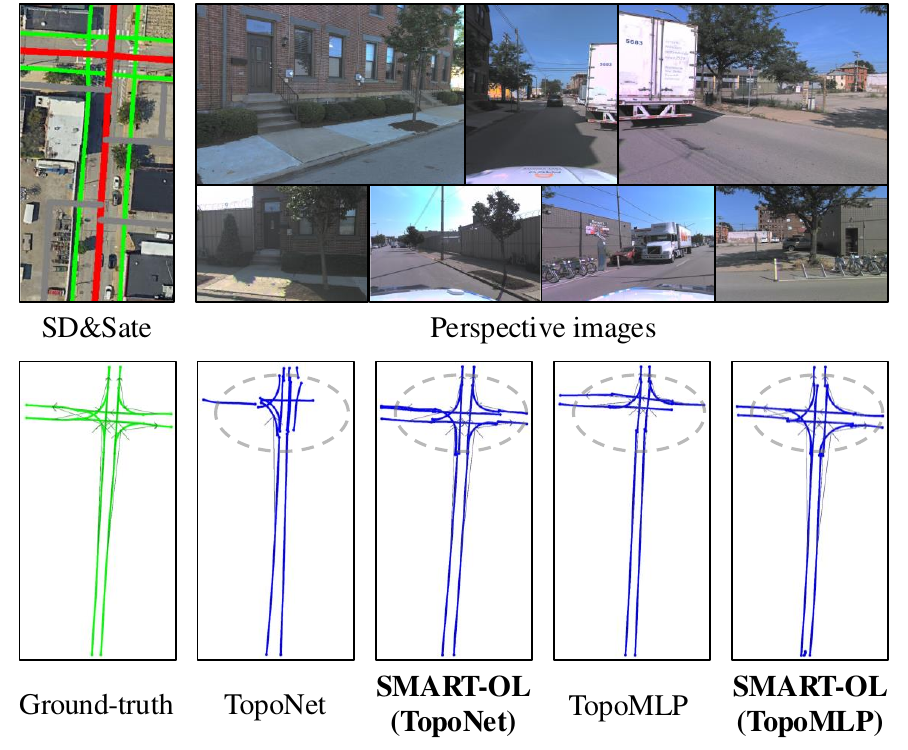}
  \vspace{-18pt}
\caption{\textbf{Qualitative comparison of \methodname-OL to baselines.} The top-left shows the SD map plotted on top of the satellite image. Our method improves baselines consistently, producing more complete lane graphs.
} 
\label{fig:vis}
\vspace{-10pt}
\end{figure}

In light of our goal to achieve robust \textit{online} driving topology reasoning, we expect \methodname also to yield state-of-the-art performance in \textit{online} settings. Therefore, we train two kinds of online topology reasoning baselines~\cite{li2023toponet, wu2024topomlp} jointly with map priors extracted by well-trained \methodname, in comparison to baselines along with state-of-the-art approaches. All models are trained for 24 epochs on the training set of~\cite{wang2023openlanev2}. As shown in Table~\ref{tab:prior_online_mapping}, \methodname-OL consistently improves baselines by over 20\% on OLS. Notably, \methodname-OL improves lane detection performances of both baselines by over 60\%, and boosts TopoNet's lane topology ability by 152.3\%. Owing to enhanced lane detection, the performance of lane-to-traffic-element topology is also improved by over 20\%. In comparison to previous works which only introduce SD maps as priors, \methodname-OL remains superior in all metrics. More specifically, TopoNet, when enhanced by \methodname, substantially outperforms both SMERF~\cite{luo2024smerf} and TopoOSMR~\cite{zhang2024enhancingonlineroadnetwork}, which also use TopoNet as their baseline. Some qualitative comparisons are illustrated in Fig.~\ref{fig:vis}, where \methodname-OL generates more complete lane graphs compared to both baselines. These results verify the efficacy of adopting map priors from \methodname to online mapping.

\Remark Since the traffic elements are detected directly from perspective images, as expected, the performance of \methodname-OL on ${\rm DET_t}$ remains similar to the baselines. 



\subsection{How well can \methodname generalize to unseen areas?}

\begin{table}[!t]\centering
    \begin{spacing}{0.8}
    \caption{\textbf{Evaluation in unseen areas.} \methodname-OL improves the baseline on all metrics, especially on ${\rm DET_l}$ and $\rm TOP_{ll}$.}\label{tab:prior-geo-disjoint}
    \end{spacing}
    \footnotesize
        \begin{tabular}{ccccccc}\toprule
        Methods &$\rm DET_l$ &$\rm TOP_{ll}$ & $\rm DET_t$ &$\rm TOP_{lt}$ & \cellcolor{gray!30}OLS \\\cmidrule{1-6}
        Baseline &16.6 &4.7 &32.9 &13.3 &\cellcolor{gray!30}26.9 \\
        \textbf{\methodname-OL} &\textbf{24.7} &\textbf{11.7} &\textbf{34.7} &\textbf{16.7} &\cellcolor{gray!30}\textbf{33.6} \\\cmidrule{1-6}
        Improvements (\%) &48.8 &148.9 &5.5 &25.6 &\cellcolor{gray!30}24.9 \\
        \bottomrule
        \end{tabular}
\end{table}

Previous works have identified that OpenLane-V2~\cite{wang2023openlanev2} contains geographically overlapping areas between training and validation sets~\cite{lilja2024localizationevaluatedataleakage, luo2024smerf}. To evaluate the performance of SMART compared to the baseline in completely unseen areas, we resplit the training and validation sets in~\cite{wang2023openlanev2}, along with the data used for the first-stage training, ensuring geo-disjoint training and evaluation. The performance comparison on the geo-disjoint splits is shown in Table~\ref{tab:prior-geo-disjoint}. Notably, \methodname-OL improves the baseline in terms of lane detection by \textbf{48.8}\% and lane topology reasoning by \textbf{148.9}\%, yielding the OLS of 33.6, significantly improving baseline's generalizability to novel areas.


\subsection{Can \methodname benefit from scaled training data?}


\begin{table}[!t]\centering
    \begin{spacing}{0.8}
    \caption{\textbf{Scaling studies for data size.} The performances of \methodname in both \textit{offline} and \textit{online} settings improve progressively as the amount of training data grows. 
    }\label{tab:scale}
    \end{spacing}
    \footnotesize
    \resizebox{\linewidth}{!}{%
        \begin{tabular}{c|cc|cccccc}\toprule
            \multirow{2}{*}{Data amount} &\multicolumn{2}{c|}{Offline} &\multicolumn{5}{c}{Online} \\\cmidrule{2-8}
            &${\rm DET_l}$ &${\rm TOP_{ll}}$ &${\rm DET_l}$ &${\rm TOP_{ll}}$ &${\rm DET_t}$ &${\rm TOP_{lt}}$ &\cellcolor{gray!30} OLS \\\cmidrule{1-8}
            Base &33.0 &16.2 &34.8 &20.9 &47.9 &27.7 &\cellcolor{gray!30}45.3 \\
            18$\times$ &34.5 &28.9 &43.8 &24.0 &48.0 &31.6 &\cellcolor{gray!30}49.3 \\
            \textbf{40$\times$} &\textbf{37.9} &\textbf{31.9} &\textbf{46.1} &\textbf{27.5} &\textbf{48.3} &\textbf{33.1} &\cellcolor{gray!30}\textbf{51.1} \\
            \bottomrule
        \end{tabular}
    }
\vspace{-8pt}
\end{table}


Scaling law has been found in Transformer-based models, and scaling the data size generally leads to improved performance. In Table~\ref{tab:scale}, we experiment \methodname with different amounts of training data, where \textit{Base} corresponds to the number of HD maps in the training set of sensor dataset~\cite{Argoverse2}. With the increases in data size to 18$\times$ and 40$\times$ larger, the performance of \methodname in offline settings increases progressively. 
We also observe significant improvements in lane topology reasoning as data size increases, growing from 16.2 to 28.9 and 31.9 when trained with 18$\times$ and 40$\times$ larger data size, respectively. Improved performance in offline settings is consistent in online settings when enhancing the online model with the corresponding \methodname trained offline. This scaling property verifies \methodname's great potential for generalizable topology reasoning in complex scenarios.


\subsection{Can SD and satellite map fusion boost performance?}

We empirically ablate SD or satellite maps to study the importance of each modality. As shown in Table~\ref{tab:ablation_sd_satellite}, the elimination of either modality leads to a degradation in both offline and online settings, with the removal of satellite maps causing a more significant decline in performance. This observation suggests that lane-level textures contained in satellite images are crucial for topology reasoning. The superior performance yielded by fusing both modalities indicates that the structural information from SD maps and the semantics from satellite images complement each other.  

\subsection{Can \methodname reduce the reliance on sensor data?}

\begin{table}[!t]\centering
    \begin{spacing}{0.8}
    \caption{\textbf{Ablation study on SD and satellite maps.} The removal of either modality leads to a degradation in performance, indicating the necessity of fusing both.
    }\label{tab:ablation_sd_satellite}
    \end{spacing}
    \footnotesize
    \resizebox{\linewidth}{!}{%
        \begin{tabular}{c|cc|cccccc}\toprule
            \multirow{2}{*}{Prior type} &\multicolumn{2}{c|}{Offline} &\multicolumn{5}{c}{Online} \\\cmidrule{2-8}
            &${\rm DET_l}$ &${\rm TOP_{ll}}$ &${\rm DET_l}$ &${\rm TOP_{ll}}$ &${\rm DET_t}$ &${\rm TOP_{lt}}$ &\cellcolor{gray!30}OLS \\\cmidrule{1-8}
            SD maps &24.7 &12.4 &34.2 &16.5 &\textbf{48.9} &27.8 &\cellcolor{gray!30}44.1 \\
            Satellite maps &35.1 &23.4 &41.2 &22.2 &48.7 &30.7 &\cellcolor{gray!30}48.1 \\\cmidrule{1-8}
            \textbf{SD and satellite maps} &\textbf{37.9} &\textbf{31.9} &\textbf{46.1} &\textbf{27.5} &48.3 &\textbf{33.1} &\cellcolor{gray!30}\textbf{51.1} \\
            \bottomrule
        \end{tabular}
    }
    \vspace{-6pt}
\end{table}

\begin{figure}[!t]
\centering
  \includegraphics[width=0.99\linewidth]{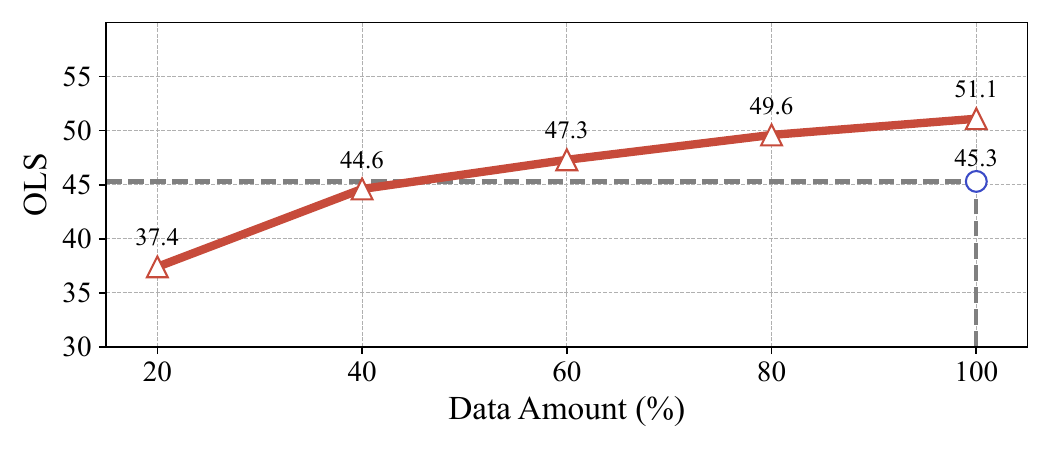}
  \vspace{-15pt}
\caption{
\textbf{Impact of varying sensor data availability.}
With only 40\% of sensor data, \methodname-OL achieves performance comparable to using the full sensor data, demonstrating its robustness with reduced sensor data.
} 
\label{fig:reduce_data}
\vspace{-10pt}
\end{figure}

\methodname trained on massive SD and satellite maps exhibits strong performance of topology reasoning. 
To investigate whether scaling up \methodname can compensate for limited sensor data in online mapping, we compare against the baseline online mapping model enhanced with \methodname trained on geospatial maps corresponding to \textit{sensor} set. We then progressively reduce the amount of sensor data used to train the online mapping model enhanced by \methodname trained on \textit{full} geospatial maps. As shown in Fig.~\ref{fig:reduce_data}, even with just 40\% of sensor data, the model utilizing priors generated from \textit{full} \methodname achieves performance comparable to that of the model using priors from \textit{sensor} \methodname, which adopts the complete sensor dataset for training.
This suggests that scaling up \methodname can significantly enhance online mapping performance, even with reduced sensor data.



\section{Conclusion and Future Works}
This paper introduces \methodname, offering a new perspective on scalable and generalizable driving topology reasoning while circumventing the need for extensive sensor data. By leveraging readily available geospatial maps and existing large-scale HD map datasets, \methodname yields impressive offline topology reasoning and supplies robust map prior representations that can be seamlessly integrated into any online driving topology reasoning architectures, achieving state-of-the-art performance.
More broadly, \methodname opens up promising avenues for future research: (1)~scaling up \methodname in both model size and data to develop a comprehensive map foundation model, and (2)~exploring the immense potential of map prior features for other tasks, such as trajectory prediction, motion planning, and end-to-end driving, wherever a robust understanding of lane structures is critical. We strongly believe that this work will considerably advance scalable and generalizable driving topology reasoning in autonomous driving.


\section*{ACKNOWLEDGMENT}
We'd like to acknowledge our friends and colleagues, including Katie Z Luo, Cheng Zhao, Arun Das, Pranav Ganti, Nikhil Advani, and Sarthak Gupta, for their fruitful discussions and follow-ups.

\bibliographystyle{IEEEtran}
\bibliography{icra25}

\end{document}

%% file: tab/online_comparison.tex
\begin{table*}[!t]\centering
    \begin{spacing}{0.8}
    \caption{\textbf{Overall performance comparison on the OpenLane-V2 Dataset.} Integrating map prior features into online mapping pipelines, \methodname-OL consistently boosts different kinds of online mapping pipelines by a wide margin, yielding state-of-the-art online driving topology reasoning performance.
    }\label{tab:prior_online_mapping}
    \end{spacing}
    \footnotesize
    \begin{tabular}{ccccccccc}\toprule
        Input type &Venues &Methods &${\rm DET_l}$ &${\rm TOP_{ll}}$ &${\rm DET_t}$ &${\rm TOP_{lt}}$ &\cellcolor{gray!30} OLS \\\cmidrule{1-8}
        \multirow{8}{*}{Perspective images} &ICCV 2021 &STSU~\cite{Can2021STSU} &12.7 &2.9 &43.0 &19.8 &\cellcolor{gray!30}29.3 \\ 
        &ICML 2023 &VectorMapNet~\cite{liu2023vectormapnet} &11.1 &2.7 &41.7 &9.2 &\cellcolor{gray!30}24.9 \\ 
        &ICLR 2023 &MapTR~\cite{MapTR} &17.7 &5.9 &43.5 &15.1 &\cellcolor{gray!30}31.0 \\ 
        &Arxiv 2023 &TopoNet~\cite{li2023toponet} &28.6 &10.9 &48.6 &23.8 &\cellcolor{gray!30}39.8 \\ 
        &Arxiv 2024 &TopoLogic~\cite{fu2024topologic} &29.9 &23.9 &47.2 &25.4 &\cellcolor{gray!30}44.1 \\ 
        &Arxiv 2024 &Topo2D~\cite{li2024topo2d} &29.1 &22.3 &\textbf{50.6} &26.2 &\cellcolor{gray!30}44.5 \\ 
        &ICLR 2024 &TopoMLP~\cite{wu2024topomlp} &28.5 &21.7 &\underline{49.5} &26.9 &\cellcolor{gray!30}44.1 \\ 
        &ECCV 2024 &RoadPainter~\cite{ma2024roadpainter} &30.7 &22.8 &47.7 &27.2 &\cellcolor{gray!30}44.6 \\\cmidrule{1-8}
        \multirow{4}{*}{Perspective images + SD maps}  
        &IROS 2024 &TopoOSMR~\cite{zhang2024enhancingonlineroadnetwork} &30.6 &17.1 &44.6 &26.8 &\cellcolor{gray!30}42.1 \\ &ICRA 2024 &SMERF~\cite{luo2024smerf} &33.4 &15.4 &48.6 &25.4 &\cellcolor{gray!30}42.9 \\
        &Arxiv 2024 &TopoLogic~\cite{fu2024topologic} &34.4 &28.9 &48.3 &28.7 &\cellcolor{gray!30}47.5 \\ 
        &ECCV 2024 &RoadPainter~\cite{ma2024roadpainter} &36.9 &\underline{29.6} &47.1 &29.5 &\cellcolor{gray!30}48.2 \\\cmidrule{1-8}
        \multirow{4}{*}{Perspective images + Map priors} & \multirow{4}{*}{\textbf{Ours}} &\multirow{2}{*}{\textbf{\methodname-OL} (TopoNet)} &\underline{46.1} &27.5 &48.3 &\textbf{33.1} &\cellcolor{gray!30}\underline{51.1} \\ 
        & & & +61.2\% & +152.3\% & -0.6\% & +39.1\% & \cellcolor{gray!30}+28.4\% \\
        & &\multirow{2}{*}{\textbf{\methodname-OL} (TopoMLP)} &\textbf{46.6} &\textbf{37.0} &47.7 &\underline{33.0} &\cellcolor{gray!30}\textbf{53.1} \\
        & & & +63.5\% & +70.5\% & -3.6\% & +22.7\% & \cellcolor{gray!30}+20.4\% \\
        \bottomrule
    \end{tabular}
    \vspace{-6pt}
\end{table*}
